\documentclass{article} 
\usepackage{nips13submit_e,times}
\usepackage{hyperref}
\usepackage{url}
\usepackage[pdftex]{graphicx}
\usepackage{amssymb,amsfonts,amsmath,amsthm}

\usepackage{fancyhdr}

\newcommand{\comment}[1]{}

\newcommand{\argmin}{\operatornamewithlimits{argmin}}
\newcommand{\argmax}{\operatornamewithlimits{argmax}}

\title{Bounded Rational Decision-Making in Changing Environments}

\author{
Jordi Grau-Moya \\
Max Planck Institute for Intelligent Systems\\
Max Planck Institute for Biolog. Cybernetics \\
T\"ubingen, Germany  \\
\texttt{jordi.grau@tuebingen.mpg.de} \\
\And
Daniel A. Braun \\
Max Planck Institute for Intelligent Systems\\
Max Planck Institute for Biolog. Cybernetics \\
T\"ubingen, Germany  \\
\texttt{daniel.braun@tuebinen.mpg.de} \\
}

\nipsfinalcopy 

\begin{document}

\maketitle

\begin{abstract}
A perfectly rational decision-maker  chooses the best action with the highest utility gain from a set of possible actions. The optimality principles that describe such decision processes do not take into account the computational costs of finding the optimal action.  Bounded rational decision-making addresses this problem by  specifically trading off information-processing costs and expected utility.  Interestingly, a similar trade-off between energy and entropy arises when describing changes in thermodynamic systems. This similarity has been recently used to describe bounded rational agents. Crucially, this framework assumes that the environment does not change while the decision-maker is computing the optimal policy. When this requirement is not fulfilled, the decision-maker will suffer inefficiencies in utility, that arise because the current policy is optimal for an environment in  the past. Here we borrow concepts from non-equilibrium thermodynamics to quantify these inefficiencies and illustrate with simulations its relationship with computational resources. 
\end{abstract}

\maketitle

\section{Introduction}

Classical decision-making theories assume that a perfect rational decision-maker should always pick the option with the best expected utility, thus ignoring the computational costs that the search for the best option entails. Experiments in decision-making under uncertainty have been shown to violate these classical theories~\cite{Ellsberg1961}. As a consequence many alternative explanations have been proposed to explain, at first sight, this irrational behaviour~\cite{guy2012decision,Ortega2013,ortegabraun2011,rubin2012trading}. A recent theory of bounded rationality for decision-making has been proposed that takes into account the computational cost of the search of the optimal policy.  Similar ideas are getting increasing interest and being used in different fields as control , robotics and machine learning ~\cite{kappen:149,kappen2005,Todorov14072009,peters2010relative,5967366}.

The theory of bounded rationality shares the same mathematical framework used in statistical physics  to describe changes in thermodynamic systems~\cite{Ortega2013,wolpert2006information,still2012thermodynamics}. In the same way a thermodynamic system trades-off its internal energy with an entropic cost---that is higher for high temperature---a bounded rational agent trades-off the expected utility of a computed policy with a  information-processing cost---that is higher for low rationality. 
Furthermore, bounded rationality takes into account model uncertainty, meaning that the model used to describe the world or policy is wrong or partially incorrect, thus allowing deviations from this model~\cite{Grau-Moya06102013,grau-moya2012}. 

However, the theory assumes that the utility function  shown to the agent  remains unchanged for the time that the agent spends computing until he samples his optimal action. An interesting problem arises when the agent is facing changing environments in such a way that the he cannot compute the optimal policy instantaneously but he still can use the previously computed policy to make a decision. In the present paper we look at the inefficiencies in utility gains due to non-optimal policy changes. In particular, we consider the special case where the used policy lags behind the optimal policy. This could be useful to explain human decision-making in fast changing environments with time-scales close to reaction times but also it serves as a first step towards having a measure of inefficiency of mechanistic systems that can use computational resources and they have to allocate them optimally.

\section{Decision-making with information-processing costs}

The deliberation process of a bounded rational decision maker consists in the transformation of a prior probabilistic model or policy $p_0(x)$ of an action $x \in \mathcal{X}$ into a posterior policy $p_1(x)$ taking into account the utility function $\Delta U(x)$ and the transformation cost from the prior distribution to the posterior distribution. This trade-off is characterized by the \textit{free energy difference (FED)}\footnote{  Our free energy difference corresponds, in statistical physics, to the negative free energy difference because of the use of energies (costs) instead of utilities.} \cite{Ortega2013}.

\begin{equation}\label{DeltaF[p]}
\Delta F [p]= \underbrace{\sum_x p(x)\Delta U(x)}_{\text{Expected utility under } p} \underbrace{- \frac{1}{\beta} \sum_x p(x)\log \frac{p(x)}{p_0(x)}}_{\text{Transformation cost}}.
\end{equation}  

The first term is the expected utility under policy $p$ and the second term is a transformation cost measured by the Kullback-Leibler  divergence ($D_{\text{KL}}(p||p_0)=\sum_x p(x)\log \frac{p(x)}{p_0(x)}$) between an arbitrary distribution $p$ and the equilibrium distribution $p_0$. The resource parameter $\beta$ sets the relative importance between the transformation cost and the maximization of $\Delta U(x)$. The optimal policy $p_1(x)$ for $\beta >0$ can be found by the maximization of the \textit{FED} that is $p_1(x)= \argmax_{p} \Delta F [p]$. However, for $\beta < 0$, $\frac{1}{\beta}D_{\text{KL}}(p||p_0)$ is concave and thus the optimal policy is found  by minimizing the \textit{FED} that is $ p_1(x)= \argmin_{p} \Delta F [p] $. For $\beta=0$, $p_1(x)=p_0(x)$. The interpretation of positive  $\beta$ is that the agent finds himself in a collaborative or exploitable environment, whereas for negative $\beta$ he finds himself in an adversarial environment with presumed rationality $\beta$.
  The solution for positive or negative $\beta$ is the same:

\begin{equation}\label{eq:posterior}
 p_1(x)= \frac{1}{Z} p_0(x) e^{\beta \Delta U(x)} 
\end{equation}

with partition function $ Z=\sum_x p_0(x) e^{\beta \Delta U(x)} $.

Replacing the non-optimal policy $p$ for the optimal policy $p_1$ in Equation~\ref{DeltaF[p]} the free energy difference becomes

\begin{equation}\label{eq:FEforward}
 \Delta F= \sum_x p_1(x)\Delta U(x) - \frac{1}{\beta} \sum_x p_1(x)\log \frac{p_1(x)}{p_0(x)}=\frac{1}{\beta}\log Z
\end{equation}

It will be useful in later parts of the paper to re-express the optimal posterior distribution of Equation~\ref{eq:posterior} in the following terms:

\begin{equation}\label{eq:posteriorReduced}
p_1(x)=p_0(x)e^{\beta \Delta U(x) - \beta \Delta F}
\end{equation}

and 

\begin{equation}\label{eq:DeltaU}
\Delta U(x)  = \frac{1}{\beta}\log \frac{p_1(x)}{p_0(x)} + \Delta F
\end{equation}

In summary, the process of decision-making proceeds as follows. At the beginning the agent finds himself in an environment where he is optimal given his computational constraints with the policy $p_0$. Then he experiences a change in the environment, and thus he observes a change in the utility function measured by $\Delta U(x)$. Given this change in the environment the previous policy $p_0$ is not optimal anymore and then he needs to compute the new optimal policy $p_1$ given the resources $\beta$. In order to do so, the process of computation requires the maximization of the free energy of Equation~\ref{DeltaF[p]}. An assumption of this process is that  if the environment changes instantaneously, the change in utility $\Delta U(x)$ has to remain the same during the whole computation and it can only change again when the agent has sampled an action from the already computed optimal policy. 

Importantly, if this assumption is not fulfilled the agent is not applying the optimal policy. In the next section we are going to give the thermodynamic interpretation of the above framework for decision-making, and moreover we are going to go further by looking at non-equilibrium thermodynamics that will serve as a motivation for the description of the aforementioned  inefficiencies in decision-making with information-processing constraints.

\section{Non-equilibrium thermodynamics}

Time evolution of a thermodynamic system can be described by a trajectory in  phase space that specifies the position and velocity of all particles at any given time.  When some external parameters $\lambda$ of such a system vary with time (e.g. a change in a magnetic potential or position of the wall of a piston) work is being applied to the system and it will evolve along a path from an initial state $A$ to a final state $B$. When the change in the parameters is done infinitely slowly  the ensemble of microstates at any specified time can be described by an equilibrium distribution  and  the work performed on such a system is equivalent to the free energy difference $W=\Delta F=F_B - F_A$.  However, when the parameters are applied in finite time, the work performed on the system will depend on the initial microscopic conditions and will be on average higher than the free energy difference \cite{jarzynski2013equalities}
\begin{equation}\label{eq:SLinequality}
\overline{W} \ge \Delta F.
\end{equation}
 
 In this case, when changing $\lambda$ the ensemble of microstates cannot be described by an equilibrium distribution and then it is said that the system is in non-equilibrium. 

We can define  the equilibrium distribution when the external parameter is held fixed.
Without loss of generality, if the parameter $\lambda \in [0,1]$ controls a switching process between an initial potential\footnote{A potential in physics can be thought as a negative utility function in decision-making} $U_A$ ($\lambda=0$) and  the final potential $U_B$ ($\lambda=1$), the equilibrium distribution for any intermediate $\lambda$ can be described with the Boltzmann distribution

\begin{equation}\label{eq:plambda}
 p_{\lambda}(x)=\frac{1}{Z_{\lambda}}e^{-\beta U_{\lambda}(x)}
\end{equation}

where $U_{\lambda}(x)=U_A(x) + \lambda (U_B(x) -U_A(x))$. In statistical physics, it is well known that the Boltzmann distribution solves the variational problem of minimizing the free energy $F=U-TS$ that is a trade-off between the internal energy of the system and an entropic cost times a temperature. More formally, for any given external parameter $\lambda$ the Boltzmann distribution satisfies: 

\[ p_{\lambda}(x)= \argmin_{p} F_{\lambda}[p]= \argmin_{p} \quad \sum_x p(x)U_{\lambda}(x) + \frac{1}{\beta} \sum_x p(x) \log p(x) \]

where $\beta=1/kT$, $k$ is the Boltzmann constant and $T$ the temperature. Note that the free energy can also be expressed as $F_{\lambda}=-\frac{1}{\beta}\log Z_{\lambda}$ and thus it is only defined in equilibrium states, so  for an arbitrary $p$, $ F_{\lambda}[p]$  is in fact a non-equilibrium free energy.

Importantly, when the switching is produced in finite time, the system will find itself in a non-equilibrium state that cannot be described by Equation~\ref{eq:plambda}. Furthermore, the work applied to the system is going to be higher than the free energy difference. 
The average \emph{extra} work applied to the system is 
 
 \[ W_{diss}=\overline{W} - \Delta F.\] 
 
When the switching process ends at $\lambda=1$  the system will start to equilibrate towards the equilibrium distribution $p_B$ under the potential $U_B$. During this process the extra work $W_{diss}$ will be dissipated in form of heat to the environment at temperature $T$ and the non-equilibrium  free energy difference  $\Delta F[p]:= F_B[p] - F[p_A]$ of the system will be minimized towards the equilibrium free energy difference $\Delta F$, assuming that the system was initially in equilibrium. 
The dissipated work, in other words, is a measure of the inefficiency of a process that drives the system from an equilibrium state $A$ to another equilibrium state $B$.

Recent advances in non-equilibrium thermodynamics have shown  a remarkable fact, that is  transforming the inequality of Equation~\ref{eq:SLinequality} into an equality (called Jarzynski equality \cite{jarzynski1997}):

\begin{equation}\label{eq:jarzynski}
e^{-\beta \Delta F}=\overline{e^{-\beta W}}
\end{equation}

where the over-line denotes an average over all possible realizations of a process that drives the system from an equilibrium state $A$ to, in general, a non-equilibrium state $B$, and $W$ denotes the work spent in such a process. 
Specifically, the above equality says that, no matter how the driving process is implemented, we can specify equilibrium quantities  from work fluctuations in the non-equilibrium process. Or in other words, this equality connects non-equilibrium thermodynamics with equilibrium thermodynamics.  
We will borrow the previous results to describe inefficiencies in the process of decision-making in the next section.

\section{Inefficiencies due to lag in the policy}

In this section we are going to use the aforementioned non-equilibrium results from statistical physics to describe inefficiencies due to not using an optimal policy. We are going to consider two simple scenarios where the change in utility function $\Delta U(x)$ is applied instantaneously or in $N$ timesteps. In both cases, we assume that the agent takes one timestep to notice the change in utility function, then samples an action from the previous policy and finally computes the optimal policy instantaneously. Due to this lag in the optimal policy the amount of expected utility gained by the agent will not be the optimal one (where the optimal one is the free energy difference) thus having inefficiencies. We will quantify these inefficiencies similarly to the dissipated work in non-equilibrium thermodynamics.

\subsection{One-step scenario}

In the one-step scenario the agent has to sample only one action. At the beginning of the process he is using an initial policy $p_0(x)$ that is optimal for the utility function $U_0(x)$ given his resources $\beta$. Then there is an external change  switching the utility function  from   $U_0(x)$ to $U_1(x)$ such that it provokes a  $\Delta U(x)$ from the point of view of the agent. Importantly, at the moment of  this increase in the utility function he is still using his previous policy $p_0$. This process is described with the following table:

\begin{center}
\begin{tabular}{l | c c}
Timestep $t$ & $0$ & $1$ \\ \hline
Utility function & $U_0(x) $ & $U_1(x)$\\
Policy & $p_0(x)$ & $p_0(x)$
\end{tabular}
\end{center}

The  average in expected utility difference from  $t=0$ to  $t=1$ is 
 
 \[ \overline{U_{net}}= \sum_x p_0(x)\Delta U(x)\]  

This quantity is the average net utility that the agent gains using the non-optimal policy. 
The average dissipated or ``wasted'' utility because not using the  optimal policy is the difference between the optimal increase in utility $\Delta F$ and the net utility  $\overline{U_{net}}$:

\begin{align}
\overline{U_{diss}}  &:=   \Delta F - \overline{U_{net}} \label{eq:Unetdiss} \\
& =   \Delta F -\sum_x p_0(x)\Delta U(x) \nonumber\\
& = \Delta F - \sum_x p_0(x)\left[ \frac{1}{\beta}\log \frac{p_1(x)}{p_0(x)} + \Delta F \right] \label{eqstep:DeltaU}\\
& = \frac{1}{\beta}\sum_x p_0(x) \log \frac{p_0(x)}{p_1(x)} \label{eq:U_diss}.
\end{align}

Equation~\ref{eqstep:DeltaU} is obtained by using Equation~\ref{eq:DeltaU} and the step from \ref{eqstep:DeltaU} to \ref{eq:U_diss} is done noticing that $\Delta F$ is a constant under the expectation over $p_0$  and cancels out.

Apart from the description of the inefficiencies derived above, the Jarzysnki-like equality for decision making can be recovered in the following way.
From Equation~\ref{eq:posteriorReduced} we can re-arrange the terms to have:

\begin{equation*}
\frac{p_1(x)}{p_0(x)}e^{\beta \Delta F}=  e^{\beta \Delta U(x)}
\end{equation*}

 Doing an expectation over the initial conditions $p_0$ yields

\begin{align*}
\sum_x p_0(x) \frac{p_1(x)}{p_0(x)}e^{\beta \Delta F}= & \sum_x p_0(x) e^{\beta \Delta U(x)}
\end{align*}

where the left term is a constant under the expectation and the right term is actually the average over  all possible  realizations of the process, thus giving:

\begin{align*}
e^{\beta \Delta F}= & \overline{e^{\beta \Delta U(x)}} 
\end{align*}

The interpretation of this result in decision-making is that the utility gains along  the path of actions taken by an agent,  gives us information about the optimal utility gains given his computational resources.

\subsection{N-step scenario}

Consider now that the agent is exposed to the same increase in utility $\Delta U(x)$, but in $N$  timesteps. After every timestep the agent is able to notice the increase in the utility function $\frac{\Delta U(x)}{N}$ but he is still using the previous policy. Next, he computes the optimal policy for this increase in utility. The following table describes this process:

\begin{center}
\begin{tabular}{l | c c c c c c c}
Timestep & $0$ & $1$ & $ 2$ & ... & $t$ & ...& $N$\\ \hline
Utility function & $U_0(x) $ & $U_1(x)$ & $U_2(x)$ &...& $U_t(x)$ & ...&$U_N(x)$ \\
Policy & $p_0(x)$ & $p_0(x)$ & $p_1(x)$ & ...& $p_{t-1}(x)$ & ... & $p_{N-1}(x)$
\end{tabular}
\end{center}

where now $U_t(x)=U_0(x) + \frac{t}{N} \Delta U(x) $ for $t \in \mathbb{N} : 0 \leq t \leq N $ and the optimal policy at timestep $t$ builds on the previous policy thus yielding:
\begin{equation}
p_t(x)=\frac{p_{t-1}(x)e^{ \frac{\beta}{N}\Delta U(x)}}{\sum_{x'} p_{t-1}(x')e^{ \frac{\beta}{N}\Delta U(x')}}
\end{equation} 
for $t>0$. The dissipated utility at  timestep $t>0$ is 

\begin{equation}
\overline{ U_{diss}}(t)  = \frac{1}{\beta} \sum_x p_{t-1}(x) \log \frac{p_{t-1}(x)}{p_{t}(x)}
\end{equation}

and the overall dissipated utility for the whole process is 

\begin{equation}
\mathcal{U}_{diss}^N =\sum_{t=1}^{N} \overline{ U_{diss}}(t)  = \frac{1}{\beta}\sum_{t=1}^{N} \sum_x p_{t-1}(x) \log \frac{p_{t-1}(x)}{p_{t}(x)}
\end{equation}

Similar to Equation~\ref{eq:Unetdiss} we can define the net utility gain for the N-step scenario as follows:

\begin{equation}
\mathcal{U}_{net}^N :=\Delta F - \mathcal{U}_{diss}^N
\end{equation}

Note that the average dissipation is lower when more time-steps are used for the change in the potential \[ \overline{\mathcal{U}_{diss}^N} \ge \overline{\mathcal{U}_{diss}^{N+1}} \].

 We recover the one-step scenario for $N=1$, corresponding to an instantaneous change in utility. Similar to a quasi-static change in a thermodynamic system, for $N\rightarrow\infty$, we get an infinitely slow change in utility, thus  $\mathcal{U}_{diss}^N \rightarrow 0$ and then the net utility equals the free energy difference $\mathcal{U}_{net}^N=\Delta F$.
 
\subsection*{Jarzynski derivation}

In a N-step scenario, similarly to Equation~\ref{eq:DeltaU}, we have that:

\begin{align*}
\Delta F=& \frac{1}{N} \left[ \Delta U(x_1) + \Delta U(x_2) ... + \Delta U(x_t)+ ... + \Delta U(x_N) \right] - \\
& - \frac{1}{\beta}\log \frac{p_1(x_1)}{p_0(x_1)} - \frac{1}{\beta}\log \frac{p_2(x_2)}{p_1(x_2)}  ... - \frac{1}{\beta}\log \frac{p_t(x_t)}{p_{t-1}(x_t)}  ...  - \frac{1}{\beta}\log \frac{p_N(x_N)}{p_{N-1}(x_N)}\\
=& \frac{1}{N}\sum_t \Delta U(x_t) - \frac{1}{\beta}\sum_{t=1}^N \log \frac{p_t(x_t)}{p_{t-1}(x_t)}
\end{align*}

where the sub-index denotes the timestep. With this relationship the exponential of the free energy difference is 

\begin{align*}
\exp \left( \beta \Delta F \right) =& \exp \left(  \frac{\beta}{N}\sum_t^N \Delta U(x_t) - \sum_{t=1}^N \log \frac{p_t(x_t)}{p_{t-1}(x_t)} \right)\\
=& \exp \left(  \frac{\beta}{N}\sum_t^N \Delta U(x_t)\right) \prod_t^N \frac{p_{t-1}(x_t)}{p_t(x_t)}
\end{align*}

where $ \prod_t^N p_{t-1}(x_t)$ is the probability of the ``path" of actions and $\frac{1}{N}\sum_t^N \Delta U(x_t)$ is the utility gain along the path exactly as the Jarzynski formulation. So by doing the expectation over $\prod_t p_t(x_t)$ we have:
\[
\exp \left( \beta \Delta F \right) = \overline{\exp \left( \frac{\beta}{N}\sum_t^N \Delta U(x_t) \right)} \]

analogous to the one-step scenario.

\section{Simulations}
\label{sec:Simulations}
In the following simulations we are going to illustrate how the number of steps and the resources affect the dissipated utility. Consider the situation where the agent can choose between two possible actions $\mathbf{x} \in \{a,b \}$ and he observes a $\Delta U(\mathbf{x})=(-2, 5)$. Importantly, this change in the utility function is made in several timesteps, allowing him to recompute the optimal policy at the end of every timestep. In particular, for this simulation the total number of timesteps is set to $N=4$. The initial policy of the agent is just $p_0(\mathbf{x})=(0.5, 0.5)$. 

\begin{figure}

\begin{center}
\includegraphics[scale=0.4,trim=0 0 0 0]{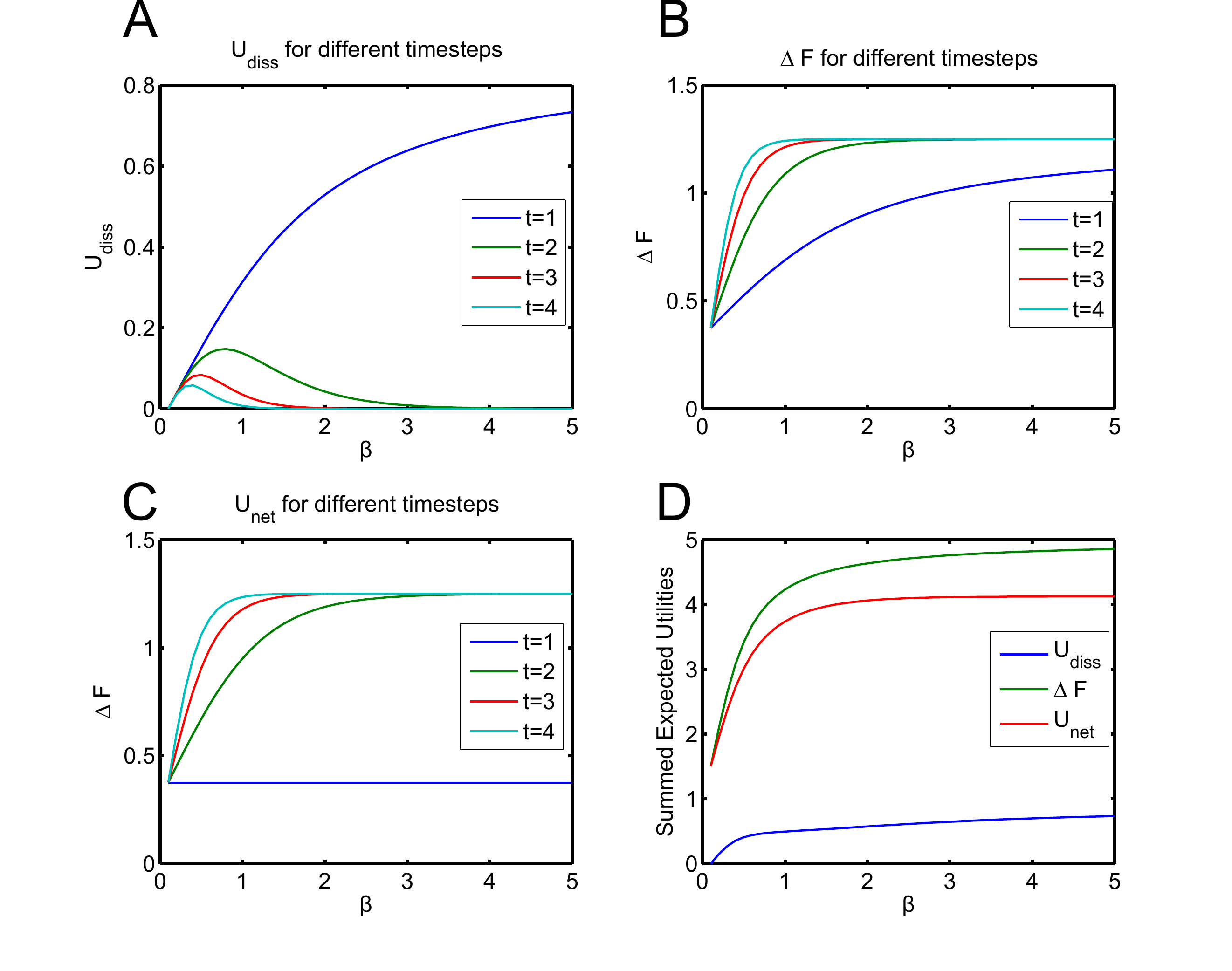}
\caption{Illustration of the dependence on $\beta $ for \textbf{A:} the dissipated utility \textbf{B:} the free energy difference \textbf{C:} the net utility \textbf{D:} the sum of the previous quantities for every timestep. See Section \ref{sec:Simulations} for a detailed description. }\label{fig:udiss}
\end{center}

\end{figure}

In this particular scenario, we show in Figure~\ref{fig:udiss}(A-C) the different values of the dissipated utility, free energy difference and the net utility with respect to the rationality parameter $\beta$. 
At the first timestep the agent is using the policy $p_0$ that does not depend on $\beta$ so the net utility is the same for all $\beta$---see Figure~\ref{fig:udiss}C. 
The increase in free energy is higher for higher $\beta$ in the different timesteps---see Figure~\ref{fig:udiss}B. In the limit case of $\beta \rightarrow 0$, the agent has no resources to change his policy and then the only gain in utility is the net utility.
The dissipated utility for the first timestep increases with increasing $\beta$ because it is a measure of inefficiency compared to the free energy difference. Notice that for this timestep, the dissipated utility is unavoidable because the agent just uses $p_0$.   However, in the next timesteps  the agent can actually use his resources to compute the optimal policy---with a lag of one timestep--- and by doing so, reduce the dissipated utility. For high $\beta=5$, the dissipated utility is almost only present in the first timestep because the agent with such a high rationality is able to quickly adopt the best policy in the second timestep, and thus being already optimal for later timesteps. This happens because we imposed a linear grow in $\Delta U(x)$ and thus best policy  for the second timestep is also the best policy for later timesteps. In general situations, this will not happen and the agent could, in principle, have high inefficiencies (high dissipated utility) even though he may have high rationality $\beta$. In Figure~\ref{fig:udiss}D we show the sum of  utilities  for \emph{all} timesteps and we see that the total net utility is less than the free energy difference plateaus for high $\beta$ only due to the inefficiencies in the first timestep.

\begin{figure}
\includegraphics[scale=0.30]{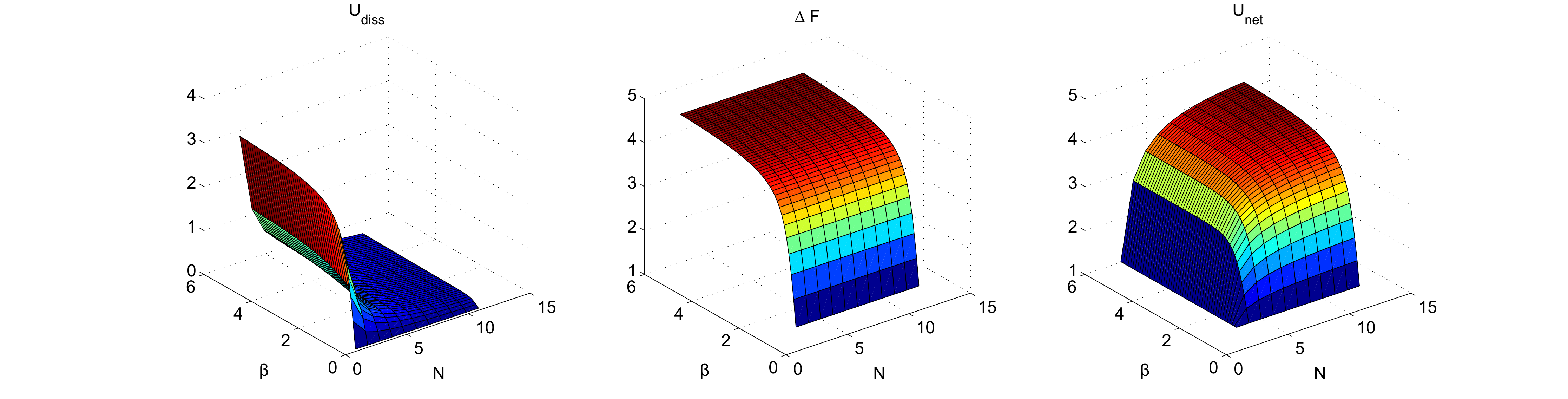}
\caption{Dependence of $\beta$ and the total number of timesteps $N$ for the sum over all timesteps of (from left to right): the dissipated utility, the free energy difference and the net utility. See Section \ref{sec:Simulations} for a detailed description. }\label{fig:color}
\end{figure}

In Figure~\ref{fig:color} we show the sum over all timesteps of the net utility, the  dissipated utility and the free energy difference for the whole process exactly as in Figure~\ref{fig:udiss}D but now also varying the total number of timesteps $N$. Note that the free energy difference is independent of $N$.   We observe that for higher $N$ the more similar the surface of net utility  is to the free energy difference. In the particular, in an infinitely slow change of utility, or in other words when $N \rightarrow \infty$, the net utility would be exactly the free energy difference. An instant switching of $\Delta U(x)$ would correspond to the case of $N=1$, where the agent is the most inefficient.

\section{Conclusions}

We described a framework of decision-making under information-processing costs and  looked at the analogies it has with the evolution of thermodynamic systems into equilibrium. We borrowed concepts from non-equilibrium thermodynamics and applied them to describe inefficiencies in decision-making due the use of non-optimal policies when the decision-maker cannot adapt perfectly to a fast changing environment. We showed an equivalent interpretation of the Jarzynski equality in thermodynamics for decision-making
that allows relating fluctuations in the possibly suboptimal achieved net utility of an agent to the optimally achievable utility given by the free energy difference. The main  contribution  of this work  is to quantify the inefficiencies that arise 
in bounded rational decision-makers when the environment changes faster than the agent can respond.  These inefficiencies  could be irrelevant in slow-changing environments but of a greater importance in fast-changing environments as  because the computed policies could differ in great deal with the optimal policies.

\subsubsection*{Acknowledgments}
This study was supported by the DFG, Emmy Noether grant BR4164/1-1.

\bibliographystyle{unsrt}
\small{
\bibliography{nips2013noneq_afternips}
}

\end{document}